\documentclass[letterpaper, 10 pt, conference]{ieeeconf}  
\usepackage{graphicx}
\usepackage{amsmath,amssymb}
\usepackage[flushleft]{threeparttable}
\usepackage{multirow}
\usepackage{xcolor}
\usepackage{amsmath}
\usepackage{amssymb}
\usepackage{cite}
\usepackage{caption}

\usepackage{tabularx}
\usepackage{booktabs}
\usepackage{bm}
\usepackage{array}
\usepackage{soul}
\usepackage[colorlinks,linkcolor=red,anchorcolor=blue,citecolor=green]{hyperref}

\IEEEoverridecommandlockouts                              

\title{\LARGE \bf
SeqTrack3D: Exploring Sequence Information for\\ Robust 3D Point Cloud Tracking
}

\author{Yu Lin, Zhiheng Li, Yubo Cui, Zheng Fang*
\thanks{This work was supported by the National Natural Science Foundation of China under Grants 62073066, the Fundamental Research Funds for Central Universities under Grant N2226001, and 111 Project under Grant B16009. (Corresponding author: Zheng Fang, e-mail: fangzheng@mail.neu.edu.cn)}
\thanks{The authors are all with the Faculty of Robot Science and Engineering, Northeastern University, Shenyang, China; 
}
}

\begin{document}

\maketitle
\thispagestyle{empty}
\pagestyle{empty}

\begin{abstract}

3D single object tracking (SOT) is an important and challenging task for the autonomous driving and mobile robotics. 
Most existing methods perform tracking between two consecutive frames while ignoring the motion patterns of the target over a series of frames, which would cause performance degradation in the scenes with sparse points. 
To break through this limitation, we introduce ``\textit{Sequence-to-Sequence}" tracking paradigm and a tracker named SeqTrack3D to capture target motion across continuous frames. 
Unlike previous methods that primarily adopted three strategies: matching two consecutive point clouds, predicting relative motion, or utilizing sequential point clouds to address feature degradation, our SeqTrack3D combines both historical point clouds and bounding box sequences. This novel method ensures robust tracking by leveraging location priors from historical boxes, even in scenes with sparse points.
Extensive experiments conducted on large-scale datasets show that SeqTrack3D achieves new state-of-the-art performances, improving by 6.00\% on NuScenes and 14.13\% on Waymo dataset. The code will be made public at \url{https://github.com/aron-lin/seqtrack3d}.

\end{abstract}

\section{INTRODUCTION}
3D Single Object Tracking (SOT) is an important task in computer vision that aims to locate a given target in subsequent frames based on its initial state. Owing to the ability to capture the continuous state changes of a specific target, 3D SOT trackers find widespread application in fields such as robotics, autonomous driving, and surveillance systems.
 
\begin{figure}[t]
	\centering
	\includegraphics[width=\linewidth]{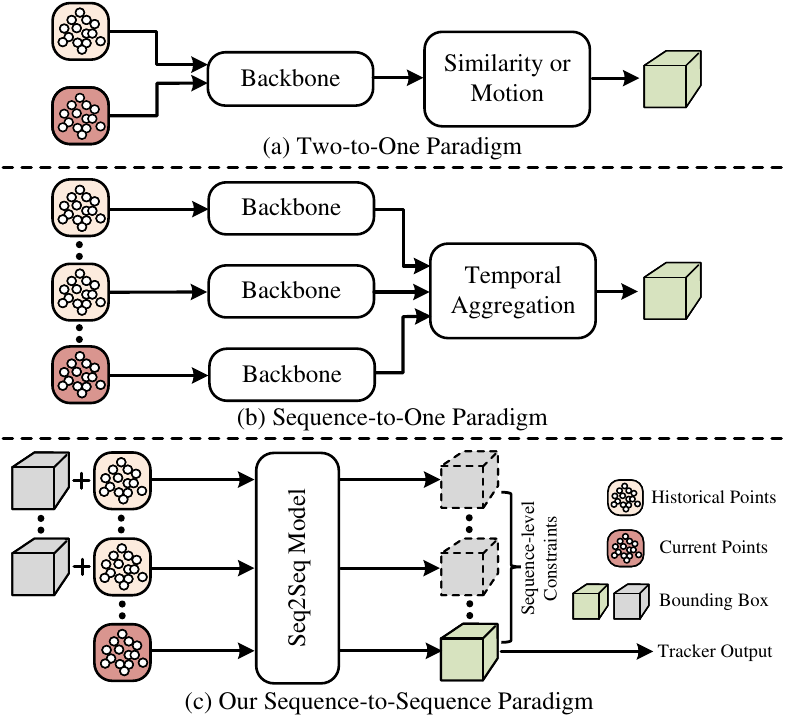}
        \caption{\textbf{The comparison of tracking paradigms.} (a) Two-to-One paradigm exploits two frames to locate target through appearance matching or motion prediction. (b) Sequence-to-One approach uses point clouds in multi-frames to integrate the target information at different times. (c) Our Sequence-to-Sequence paradigm considers temporal clues of the target in points and box sequences to overcome sparse points cases.}
	\label{fig:comparison}
\vspace{-0.2in}
\end{figure}

Inspired by image-based 2D SOT methods~\cite{SiamFC,SiamRPN,siamDW}, early 3D SOT trackers~\cite{P2B,BAT,PTT,lttr,pttr,glt-t,SC3D} usually follow the ``\textit{Two-to-One}" paradigm (in Fig.~\ref{fig:comparison}(a)), which captures the target by matching template points to a search region in the current frame. However, the sparse and textureless point clouds make it challenging to track fast-moving or occluded targets. To solve this problem, M$^2$-Track~\cite{beyond} considers tracking as a motion prediction task, estimating the relative motion between two clusters of the foreground points across frames. Nonetheless, both matching-based~\cite{P2B,BAT,PTT,lttr,pttr,glt-t,SC3D} and motion-based~\cite{beyond} methods simplify tracking to a transient prediction problem (only two frames). This not only breaks the continuity of tracking but also ignores motion pattern of target reflected in previous frames, making tracker hard to handle sparse points cases.

Lately, some works~\cite{TAT,sttrack} try to adopt a ``\textit{Sequence-to-One}" approach to merge target-specific information from historical point clouds into the current frame (in Fig.~\ref{fig:comparison}(b)). However, in the scenes with extremely sparse points or an invisible target, it is non-trivial to locate the target reliably through limited historical features. 
Compared to solely relying on point cloud sequences, we argue that the historical 3D bounding boxes (BBox) offer additional crucial insights. For instance, even when target points become sparse or are entirely absent, the current target location can still be inferred from the historical BBox trajectory which reflects the target states during a period.

Building upon the above idea, we propose a ``\textit{Sequence-to-Sequence}" (Seq2Seq) paradigm (in Fig. 1(c)), which captures the long-term motion pattern of the target in point clouds and box sequences. Specifically, our Seq2Seq framework takes a sequence of cropped points and historical BBoxes as input, performs intra- and inter-frame feature augmentation, and obtains an updated BBox sequence. The BBox in the current frame serves as the final output of tracker, while the other BBoxes enforce a sequence-level constraint to compel the network to explicitly learn the motion patterns. 

To implement Seq2Seq framework, we introduce a \textit{SeqTrack3D} tracker. It first exploits a Transformer-based encoder to extract both key geometry and motion information from the point clouds sequence. After that, the historical boxes guide the decoder to integrate target-specific information and produce embeddings for sequential BBoxes. In this way, the model can explicitly learn motion patterns of the target while projecting the point clouds sequence to the BBox sequence.
Meanwhile, instead of utilizing complex head networks such as VoteNet~\cite{votenet} in~\cite{PTT,P2B,BAT} and CenterNet~\cite{centerpoint} in~\cite{sttrack,SMAT}, we directly predict target attributes $(x, y, z, \theta)$, leading to a more efficient network.

Furthermore, we prove the effectiveness of our method on two large-scale datasets, NuScenes~\cite{NuScenes} and Waymo~\cite{waymo}. Experimental results show that SeqTrack3D outperforms the recent STTracker by 6.00\% on NuScenes and outperforms the state-of-the-art (SOTA) M$^2$-Track by 14.13\% on Waymo while running at 38 FPS.

In sammary, the contributions of our work are as follows: 
\begin{itemize}
\item We introduce a novel Seq2Seq paradigm, which advocates using sequence-level information and constraint to model continuous object motion in 3D SOT task.
\item Based on Seq2Seq paradigm, we propose a SeqTrack3D tracker that leverages an encoder-decoder structure to extract spatial-temporal information from point clouds and BBoxes sequences, resulting in remarkable performance improvements.
\item The experimental results on the NuScenes and Waymo demonstrate that our approach significantly outperforms the previous state-of-the-art methods. The code will be released to the research community.
\end{itemize}

\section{RELATED WORK}

\subsection{3D Single Object Tracking}
Early 3D SOT approaches mainly inherited siamese structure adopted in 2D SOT~\cite{SiamFC,SiamRPN,siamDW,siamcar}, which can be rigorously described as the ``\textit{Two-to-One}" paradigm. For instance, pioneering SC3D~\cite{SC3D} adopted a siamese structure to capture the target shape and generated a series of candidate boxes within the search region. The best proposal was then selected based on cosine similarity. However, heuristic sampling in SC3D hinders end-to-end training.
After that, P2B~\cite{P2B} tried to utilize VoteNet~\cite{votenet} to improve target generation mechanism and achieved end-to-end training. In subsequent studies such as PTT~\cite{PTT}, BAT~\cite{BAT}, LTTR~\cite{lttr}, SMAT~\cite{SMAT}, STNet~\cite{stnet}, etc., researchers have attempted to enhance performance through improving the quality of similarity features and optimizing target generation strategies.
Despite advances in matching-based methods~\cite{P2B,PTT,BAT}, they are limited by the sparse and incomplete natures of point clouds. To alleviate this problem, M$^2$-Track~\cite{beyond} introduced the motion-centric paradigm that located the target by estimating the relative motion between two frames and achieved remarkable performance improvement. Nevertheless, the above methods overlook that tracking is a complicated dynamic process and do not fully explore spatial-temporal contextual information.

Recently, some studies introduced the ``\textit{Sequence-to-One}" paradigm that exploits multi-frame point clouds to estimate the target box in the current. TAT~\cite{TAT} selected high-quality historical templates and aggregated historical clues through the recurrent neural networks (RNN). However, the selective usage of previous frames interrupts the continuous motion information of target. Additionally, STTracker~\cite{sttrack} encoded points sequence in the bird's eye view (BEV) space and split feature maps into sequential patches to capture target motion through deformable attention~\cite{zhu2020deformable}. Unfortunately, the target might be separated into different patches, thereby destroying appearance integrity. Thus, the approaches in \cite{TAT,sttrack} still struggle to effectively exploit historical information to locate target accurately in challenging scenes.

\begin{figure*}[t]
	\centering
	\includegraphics[width=\linewidth]{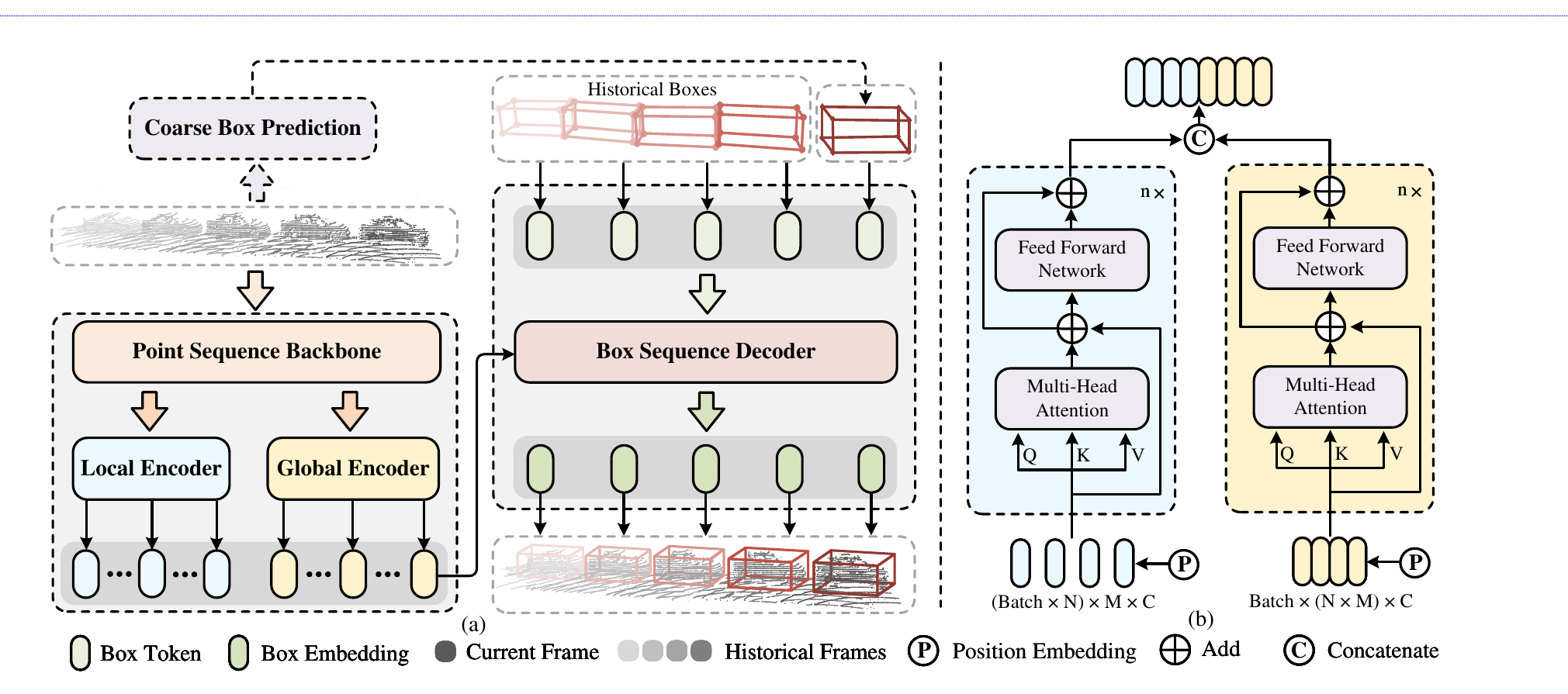}
	\caption{(a) Overview of SeqTrack3D tracker. The encoder establishes spatial-temporal relations for point sequence. Guided by object sequence with target prior, the decoder generates box embeddings and utilizes them to predict sequential bounding boxes. (b) Details of the local-global encoder that encodes point sequence in a decoupled manner.}
	\label{fig:main_figure}
\vspace{-0.5cm}
\end{figure*} 

\subsection{Transformer-based Sequence Learning} 
The Transformer structure has developed into a universal approach for sequence learning and has also been adapted for hot topics in computer vision, such as classification and object detection~\cite{VIT,swin,DETR,detr3d,segformer}. Specifically, ViT~\cite{VIT} transformed image into sequential patches and captured global contextual information by attention mechanism. Swin Transformer~\cite{swin} introduced window attention to reduce computational complexity. Meanwhile, DETR~\cite{DETR} utilized sequential patches for object detection, which has also been extended to the 3D task in DETR3D~\cite{detr3d}. Similarly, Segformer~\cite{segformer} considered segmentation as sequence learning and utilized Transformer to learn long-range dependencies among pixels. 
Inspired by the above works, we employ Transformer to construct mapping relationship between point clouds and object sequences, resulting in robust tracking.

\section{METHODOLOGY}
\subsection{Framework Overview}
3D SOT aims to identify the target bounding box \( \mathcal{B}_t = (x,y,z,w,l,h,\theta) \in \mathbb{R}^7 \) within the current point cloud \( \mathcal{P}_t \in \mathbb{R}^{W \times 3} \) based on a given initial BBox \( \mathcal{B}_{0} \). Here, \( (x, y, z) \), \( (w, l, h) \) and \( \theta \) represent the 3D center, size and orientation, respectively. \( {W \times 3} \) denotes \( W \) points with 3D coordinates.
Unlike in 2D images where target dimensions can be changed due to the different viewpoints, we assume that the size is constant in 3D space. Thus, following previous studies, we only estimate \( (x, y, z, \theta) \) in every frame.
Besides, different from approaches~\cite{sttrack,TAT} which focus on multi-frame point clouds information \( \{\mathcal{P}_{t-n}\}_{n=0}^{N} \), we additionally introduce object sequence \( \{\mathcal{B}_{t-n}\}_{n=1}^{N} \) and reformulate the tracking task as a Seq2Seq problem:
\begin{align}
\mathcal{F}\big(\{\mathcal{P}_{t-n}\}_{n=0}^{N},\{\mathcal{B}_{t-n}\}_{n=1}^{N}\big) \rightarrow \{\hat{\mathcal{B}}_{t-n}\}_{n=0}^{N}
\label{eq:paradigm}
\end{align}
where \( N \) denotes the length of time window, and \( \hat{\mathcal{B}}_{t} \) is the estimated BBox. Note that we transform both the points and BBox sequence into a unified local coordinate system with ego pose.

Based on Eq.~\ref{eq:paradigm}, we introduce a SeqTrack3D tracker that can construct spatiotemporal correlations of point sequences and exploit continuous motion cues in object sequences.
As shown in Fig.~\ref{fig:main_figure}, given a sequence \( \{\mathcal{P}_{t-n}\}_{n=0}^{N} \), SeqTrack3D first employs a weight-shared backbone to aggregate multi-frame features and predict a coarse BBox \( \mathcal{B}_{t} \), which is then used to form an object sequence \( \{\mathcal{B}_{t-n}\}_{n=0}^{N} \) with historical boxes. Next, we present a local-global encoder responsible for encoding inter- and intra-frame contextual information, thereby obtaining sequential features \( \{F_{t-n}\}_{n=0}^{N} \). Finally, the decoder utilizes \( \{F_{t-n}\}_{n=0}^{N} \) to produce final BBox sequence \( \{\hat{\mathcal{B}}_{t-n}\}_{n=0}^{N} \) with guidance of object sequence \( \{\mathcal{B}_{t-n}\}_{n=0}^{N} \).

\subsection{Coarse Box Prediction}
In SeqTrack3D, we aim to project point cloud sequences to BBox sequences. Borrowing from object detection methods such as~\cite{DETR,zhu2020deformable,detr3d}, we predefine a box query for every object at different time steps to establish query-object pairs. These queries are designed to extract target-specific features from point sequence \(\{\mathcal{P}_{t-n}\}_{n=0}^{N}\). For objects in historical frames, we adopt their historical BBoxes to generate queries. Similarly, for the current frame, we also wish to produce a query in the same approach and thus opt to estimate a coarse BBox.
For efficiency, we draw inspiration from M$^2$-Track\cite{beyond}. Firstly, we utilize a PointNet~\cite{PointNet} to generate the multi-frame point features \( H_{t-N}^{t} \in \mathbb{R}^{NW \times C} \), where \( W \) denotes the number of points in a single frame and \( C \) is the number of channels. Then, the features are fed into multi-layer perceptron (MLP) to generate a foreground point mask \( \mathcal{M} \in \mathbb{R}^{NW} \). Finally, \( H_{t-N}^{t} \) is weighted by mask \( \mathcal{M} \) and undergoes max-pooling with another MLP to obtain a coarse bounding box \( \mathcal{B}_{t} \) in the local coordinate system. The above process can be noted as:
\begin{align}
\mathcal{B}_{t} = \operatorname{MLP}\left( \operatorname{Pooling}(H_{t-N}^{t} \times \mathcal{M}) \right), \mathcal{B}_{t} \in \mathbb{R}^4
\end{align}
\subsection{Local-Global Feature Encoding}
Previous approaches like~\cite{TAT,sttrack} employ a single Transformer structure to extract the spatial and temporal features at the same time. Although the above methods are effective at capturing intricate dependencies, they could be problematic since spatial and temporal features possess distinct attributes. 
In particular, spatial features focus on relative positions and partial shapes, while temporal features reflect target motion patterns. 
Therefore, the tightly-coupled manner in~\cite{TAT, sttrack} may cause the network to be biased during optimization and overlook useful cues for target tracking. 

Inspired by video processing in~\cite{timesformer, arnab2021vivit, videotransformer}, we mitigate the above issue via ``divide-and-conquer" strategy and propose a local-global encoder that decouples the encoding of spatial and temporal features. The local encoder concentrates on perceiving target location in each frame, while global encoder constructs inter-frame relation for target dynamics over time.

In this part, we first exploit a modified PointNet++~\cite{PointNet++} as a point sequence backbone for point feature extraction and generate \( F_n \in \mathbb{R}^{M \times C} \ (n \in \{t-N,...,t \}) \). 
Contrasting with original PointNet++, our modified version with fewer set abstraction layers aims to ensure efficiency by reducing the number of point features, which also enables affordable subsequent attention calculations.
Thereafter, we feed point features \( F_n \) into local encoder for further encoding. Following 3DETR~\cite{misra2021end}, we employ LayerNorm~\cite{layernorm} to get features \( E_n \in \mathbb{R}^{M \times C} \):
\begin{align}
E_n = \operatorname{LayerNorm}(F_n)
\end{align}
Then, we project \( E_n \) into \textit{query}, \textit{key}, and \textit{value} embeddings, which is formulated as:
\begin{align}
Q_n = {E_n}{W_q}, K_n = {E_n}{W_k}, V_n = {E_n}{W_v}
\end{align}
where \( W_q, W_k, W_v \) are the linear projection matrices. After obtaining $Q_n,K_n,V_n$, we use a multi-head attention mechanism that can be described as:
\begin{align}
{head}=\operatorname{Softmax}\left(\frac{Q K^{T}}{\sqrt{{d}_k}}\right) V
\label{eq:qkv}
\end{align}
\begin{align}
\operatorname{MHA}\left(Q_n, K_n, V_n\right) = \text{Concat}\left(\mathit{head}_1,..., \mathit{head}_h\right) W^o
\label{eq:mha}
\end{align}
\begin{align}
\hat{F}_n = F_n + \operatorname{Dropout}\left(\operatorname{MHA}\left(Q_n, K_n, V_n\right)\right)
\end{align}
Here, \( h \) means the number of heads, and \( \operatorname{MHA} \) represents the attention mechanism across multiple subspaces that enhances the expressive capacity of model and increases sensitivity to different aspects of input data. Later, the features are fed into a feed-forward network (FFN) for further refinement:
\begin{align}
\tilde{F}_n = \hat{F}_n + \operatorname{Dropout}(\operatorname{FFN}(\operatorname{Layernorm}(\hat{F}_n)))
\end{align}
In the above procedure, the local encoder only calculates the attention map within each individual frame, leading to geometric features \( \{\tilde{F}_{t-N}, \ldots, \tilde{F}_{t}\} \) encoded in each frame independently. 
On the other hand, the global encoder adopts the same design as local encoder, but the difference is that we merge features $\{F_{t-N}, ..., F_{t}\}$ into $E_m \in \mathbb{R}^{NM \times C}$, which enables attention maps to span entire sequence.
Then, \( E_m \) is processed by multiple attention layers to learn inter-frame relations and embed motion features into \( \{\tilde{G}_{t-N}, \ldots, \tilde{G}_{t}\} \).

Finally, we combine the outputs from both local and global encoders to produce a sequential feature \( \tilde{F}_{gl} \in \mathbb{R}^{2MN \times C} \). In this way, the network can take into account local geometric structures and global dynamics in a balanced manner.

\subsection{Box Sequence Generating}
The intention of this section is to conduct the decoder to generate a more accurate BBox sequence via \(\{\mathcal{B}_{t-n}\}_{n=0}^{N}\) and sequential feature $\tilde{F}_{gl}$. 
To bridge the gap between BBox and point representations, we convert the BBox parameters $P = (x, y, z, w, l, h, \theta)$ into the BBox corners $C \in \mathbb{R}^{L\times K}$, where $K=(x_c, y_c, z_c, t)$ contains spatial location and timestamp, and $L$ is the number of corners. Then, we project $\{C_{t-n}\}_{n=0}^{N}$ to box token $T \in \mathbb{R}^{N \times L \times C^{\prime}}$ with location prior of target:
\begin{align}
\{C_{t-n}\}_{n=0}^N = \psi(\{{P}_{t-n}\}_{n=0}^N), \ T=\phi(\{C_{t-n}\}_{n=0}^N)
\label{eq:boxtocords}
\end{align}
where \( N, C^{\prime} \) mean the number of boxes and channel dimensions. \( \psi(\cdot) \) projects BBox parameters to BBox corners, and \( \phi(\cdot) \) performs a linear mapping. Subsequently, BBox tokens and point cloud features are transformed into \textit{query}, \textit{key}, and \textit{value} embeddings for decoder input as follows:
\begin{align}
Q' = TW_q',\ K' = \tilde{F}_{lg}W_k',\ V' = \tilde{F}_{lg}W_v' 
\label{eq:decoder}
\end{align}
Then, the decoder utilizes an attention mechanism consistent with encoder to generate BBox Embedding \( D \in \mathbb{R}^{N \times L \times C'} \), which collects target-specific information from local-global features $\tilde{F}_{gl}$. Finally, we flatten \( D \) to the size of $N \times LC'$ and feed it through the MLP to derive the sequential BBox parameters \( \{\hat{\mathcal{B}}_{t-n}\}_{n=0}^{N} \). These sequential parameters can act as constraints during training, while \( \hat{\mathcal{B}}_{t} \) becomes final output for tracking in the inference phase.

\subsection{Implementation Details}
\noindent\textbf{Loss Functions.} 
The loss function is described by Eq.~\ref{eq:loss}. Specifically, \( \mathcal{L}_{\text{mk}} \) represents the mask loss for points, which is computed using a cross entropy loss. \( \mathcal{L}_{\mathrm{cb}} \) denotes the coarse BBox loss. Both \( \mathcal{L}_{\mathrm{p}} \) and \( \mathcal{L}_{c} \) account for the loss of previous and current bounding boxes in the predicted object sequence, respectively. 
The BBox loss consists of two components: the angle loss and the center loss. Both of these components use the Huber Loss~\cite{huberloss} for computation.
To prevent the network from overly relying on historical BBox that will accumulate errors, we set the weight ratio \( \gamma_1 : \gamma_2 = 1 : 10 \).
\begin{align}
\mathcal{L} = \lambda_1 \mathcal{L}_{\mathrm{mk}} + \lambda_2 \mathcal{L}_{\mathrm{cb}} + \lambda_3 (\gamma_1 \mathcal{L}_{\mathrm{p}} + \gamma_2 \mathcal{L}_{\mathrm{c}})
\label{eq:loss}
\end{align}
\noindent\textbf{Training \& Inference.}
Our SeqTrack3D supports end-to-end training. To facilitate this, it is crucial to ensure uninterrupted gradient flow during the implementation of Eq.~\ref{eq:boxtocords}. 
Notably, we adopt random offsets to the ground truth during training to simulate the historical boxes with errors during the testing phase.
In data processing, each point cloud is attached with a timestamp and mask. The mask is set to 0 or 1 to distinguish foreground points, while the points in the current frame are assigned 0.5 as their status remains unknown.
The number of attention layers and heads $h$ in local-global encoder is set to 3 and 4, respectively. 
We train the model for 40 epochs on NVIDIA RTX 4090 GPUs using an Adam optimizer with an initial learning rate of \(1 \times 10^{-4}\).
\section{EXPERIMENTS}
\begin{table*}
  \renewcommand{\arraystretch}{1.0}
  \renewcommand\tabcolsep{8.2pt} 
  \footnotesize
  \caption{\textbf{Comparison of SeqTrack3D against state-of-the-art methods on NuScenes and Waymo Open Dataset. } }
    \vspace{-0.3cm}
    \begin{center}
    \begin{tabular}{l|c|cccccc|ccc}
        \toprule[.05cm]
        \multirow{3}*{}
        &  Dataset  & \multicolumn{6}{c|}{{NuScenes}} & \multicolumn{3}{c}{{Waymo Open Dataset}}\\ \cline{2-11} 
        \rule{0pt}{8pt}
        & Category  & Car & Pedestrian & Truck & Trailer & Bus  & Mean & Vehicle   & Pedestrian  &Mean\\
        & Frame Number &\textit{64,159} & \textit{33,227} &\textit{13,587} &\textit{3,352} &\textit{2,953} &\textit{117,278} &\textit{1,057,651} &\textit{510,533} &\textit{1,568,184} \\
        \hline \hline
        \multirow{8}*{\rotatebox{90}{Success}}
        \rule{0pt}{8pt}
        & SC3D~\cite{SC3D} & 22.31 & 11.29 & 30.67 & 35.28 & 29.35 & 20.70 & - & - &- \\
        & P2B~\cite{P2B} &38.81 & 28.39 &42.95 &48.96 &32.95 &36.48 &28.32 & 15.60 & 24.18\\
        & BAT~\cite{BAT} &40.73 & 28.83 &45.34 &52.59 &35.44 & 38.10 & 35.62 &22.05 &31.20 \\
        & GLT-T~\cite{glt-t} & 48.52 & 31.74 & 52.74 & 57.60 & 44.55 & 44.42 & - & - &- \\
        & CXTrack~\cite{xu2023cxtrack} & 48.92 & 31.67 & 51.40 & \underline{60.64} & 40.11 & 44.43  &- & - &-\\
        & PTTR~\cite{pttr} & 51.89 & 29.90 & 45.30 & 45.87 & 43.14 & 44.50 & - & - &- \\
        & M$^2$-Track~\cite{beyond} &55.85 & 32.10 &\underline{57.36} &57.61 & \underline{51.39} &49.23 &\underline{43.62} & \underline{42.10} & \underline{43.13}\\
        & STTracker~\cite{sttrack} & \underline{56.11} & \underline{37.58} &54.29 &48.13 &36.31 & \underline{49.92} & - & - & -\\
        \cline{2-11}
        \rule{0pt}{8pt}
        & \textbf{SeqTrack3D (Ours)} &\textbf{62.55}  &\textbf{39.94} &\textbf{60.97} &\textbf{68.37} & \textbf{54.33} &\textbf{55.92} &\textbf{63.56} &\textbf{44.21} &\textbf{57.26} \\
        & \textit{Improvement} &\textcolor[rgb]{0.0,0.5,0.0}{\textit{$\uparrow$6.44}} & \textcolor[rgb]{0.0,0.5,0.0}{\textit{$\uparrow$2.36}} &\textcolor[rgb]{0.0,0.5,0.0}{\textit{$\uparrow$3.61}} &\textcolor[rgb]{0.0,0.5,0.0}{\textit{$\uparrow$7.73}} &\textcolor[rgb]{0.0,0.5,0.0}{\textit{$\uparrow$2.94}} & \textcolor[rgb]{0.0,0.5,0.0}{\textit{$\uparrow$6.00}} &\textcolor[rgb]{0.0,0.5,0.0}{\textit{$\uparrow$19.94}} &\textcolor[rgb]{0.0,0.5,0.0}{\textit{$\uparrow$2.11}} &\textcolor[rgb]{0.0,0.5,0.0}{\textit{$\uparrow$14.13}} \\
        \hline \hline
        \rule{0pt}{8pt}
        \multirow{8}*{\rotatebox{90}{Precision}}
        & SC3D~\cite{SC3D} & 21.93 & 12.65 & 27.73 & 28.12 & 24.08 & 20.20 &- & - &-\\
        & P2B~\cite{P2B} &43.18 & 52.24 &41.59 &40.05 &27.41 &45.08 &35.41 & 29.56 & 33.51 \\
        & BAT~\cite{BAT} & 43.29 & 53.32 & 42.58 & 44.89 & 28.01 & 45.71 & 44.15 & 36.79 & 41.75\\
        & GLT-T~\cite{glt-t} & 54.29 & 56.49 & 51.43 & 52.01 & 40.69 & 54.33 &- & - &- \\
        & CXTrack~\cite{xu2023cxtrack} & 55.61 & 56.64 & 50.93 & 54.44 & 35.83 & 54.83  &- & - &-\\
        & PTTR~\cite{pttr} & 58.61 & 45.09 & 44.74 & 38.36 & 37.74 & 52.07  &- & - &- \\
        & M$^2$-Track~\cite{beyond} &65.09 & 60.92 &59.54 & \underline{58.26} & \underline{51.44} &62.73 & \underline{61.64} & \underline{67.31} & \underline{63.48} \\
        & STTracker~\cite{sttrack} & \underline{69.07} & \underline{68.36} & \underline{60.71} &55.40 &36.07 & \underline{66.68} & - & - & -\\
        \cline{2-11}
        \rule{0pt}{8pt}
        & \textbf{SeqTrack3D (Ours)} &\textbf{71.46} & \textbf{68.57} &\textbf{63.04} &\textbf{61.76} &\textbf{53.52} &\textbf{68.94} &\textbf{73.48} &\textbf{68.52} &\textbf{71.86}\\
        & \textit{Improvement} &\textcolor[rgb]{0.0,0.5,0.0}{\textit{$\uparrow$2.39}} & \textcolor[rgb]{0.0,0.5,0.0}{\textit{$\uparrow$0.21}} &\textcolor[rgb]{0.0,0.5,0.0}{\textit{$\uparrow$2.33}} &\textcolor[rgb]{0.0,0.5,0.0}{\textit{$\uparrow$3.57}} &\textcolor[rgb]{0.0,0.5,0.0}{\textit{$\uparrow$2.08}} & \textcolor[rgb]{0.0,0.5,0.0}{\textit{$\uparrow$2.26}} &\textcolor[rgb]{0.0,0.5,0.0}{\textit{$\uparrow$11.84}} &\textcolor[rgb]{0.0,0.5,0.0}{\textit{$\uparrow$1.21}} &\textcolor[rgb]{0.0,0.5,0.0}{\textit{$\uparrow$8.38}} \\
        \toprule[.05cm]
    \end{tabular}
    \end{center} 
    \label{tab:nusc_waymo}
    \vspace{-0.7cm}
\end{table*}
\noindent\textbf{Datasets \& Evaluation.}
We conduct extensive experiments on NuScenes and Waymo datasets. NuScenes uses 32-beam LiDAR and annotates around 1.4M instances, while Waymo employs 64-beam LiDAR. Compared with 10 Hz annotation frequency in Waymo, NuScenes is more challenging due to low fequency of only 2 Hz.
Furthermore, we utilize the One Pass Evaluation (OPE) protocol~\cite{OPE} to evaluate our SeqTrack3D through two primary metrics: \textit{Success} and \textit{Precision}. \textit{Success} quantifies the overlap between the predicted and ground-truth BBoxes via the Intersection Over Union (IOU), computing its Area Under Curve (AUC) when the overlap exceeds a threshold ranging from 0 to 1. \textit{Precision} evaluates localization accuracy of target based on the distance between BBox centers, using an AUC for distances ranging from 0 to 2 meters.

\subsection{Quantitative Results.} 
\noindent\textbf{Comparison with SOTA methods.} As displayed in Table.~\ref{tab:nusc_waymo}, we compare SeqTrack3D against previous SOTA methods, including SC3D~\cite{SC3D}, P2B~\cite{P2B}, BAT~\cite{BAT}, GLT-T~\cite{glt-t}, CXTrack~\cite{xu2023cxtrack}, PTTR~\cite{pttr}, M$^2$-Track~\cite{beyond} and STTracker~\cite{sttrack}. 
The results exhibit that our method achieves significant superiority in all categories. Specifically, on the NuScenes dataset, our method outperforms the \textit{Sequence-to-One} method~\cite{sttrack} with improvements of 6.00\%/2.26\% in \textit{Success}/\textit{Precision}. This validates the advantage of the proposed \textit{Sequence-to-Sequence} paradigm and the effectiveness of our decoupling strategy in space-time encoding. 
Furthermore, SeqTrack3D surpasses M$^2$-Track~\cite{beyond} on Waymo by 14.13\%/8.38\% with a notable increase of 19.94\%/11.84\% in the vehicle category.
Thus, we think that compared to estimating frame-to-frame motion in \cite{beyond}, our paradigm captures motion patterns over multiple frames and achieve significant performance gain.

\begin{figure}[t]
\centering
\includegraphics[width=\linewidth]{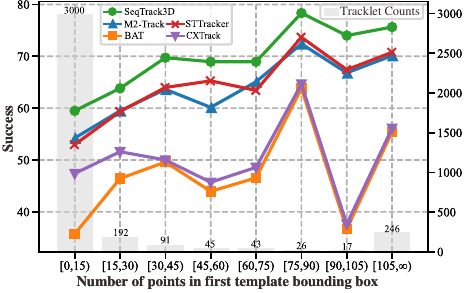}
\caption{\textbf{Tracking performance across varying numbers of template points in the first frame.}}
\label{fig:tracklet}
\vspace{-0.5cm}
\end{figure}

\noindent\textbf{Robustness to Sparsity.} In order to present a finer-grained robustness comparison, we plotted the \textit{Success} metric of different methods for the Car category under various degrees of point sparsity in Fig.~\ref{fig:tracklet}. In most sparse cases, where the initial frame contains only 0-15 points, our SeqTrack3D surpasses other methods, which validates the obvious advantage of our paradigm when dealing with sparse scenes.

\noindent\textbf{Inference Speed.} Our model is designed to extract position-related information from historical point clouds and BBoxes to model a continuous motion. Although performing feature extraction on multiple point clouds does increase the computational load, we opt for a lightweight backbone, even though a more powerful structure such as DGCNN~\cite{DGCNN} is available. As a result, our choice strikes a balance between speed and performance, enabling it to maintain competitive runtime, as shown in Table.~\ref{tab:speed}.

\subsection{Qualitative Analysis.}
\begin{figure*}[t!]
     \centering
     \includegraphics[width=\linewidth]{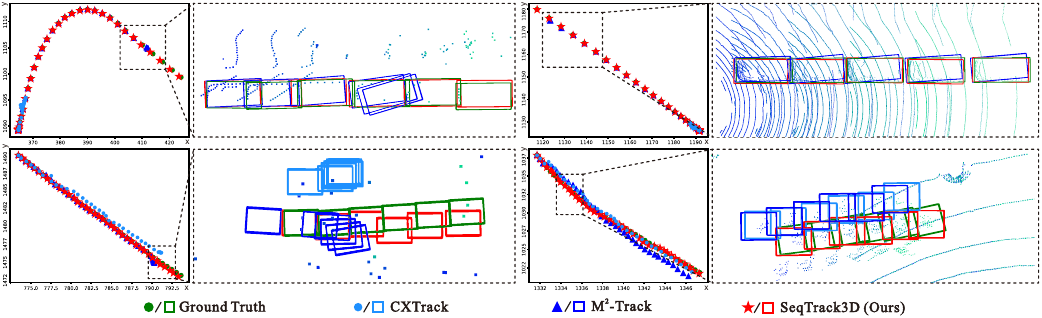}
	\caption{\textbf{Visualization of tracking results on NuScenes.} Complete tracking trajectories are projected onto the X-Y plane of the global coordinate system. Left: Point sparsity cases; Right: Dense cases; Top: Car category; Bottom: Pedestrian category.}     \label{fig:vis}
     \vspace{-0.3cm}
\end{figure*}

To better understand the behaviour of our model across different scenes, we illustrate the complete tracking trajectories in Fig.~\ref{fig:vis}. In the case of sparse points (on the left), CXTrack is hard to extract robust point features and results in early localization failure, while M$^2$-Track depend on the relative motion prediction to persist for a while. 
Despite this, once the target is continuously occluded and invisible, M$^2$-Track also tends to fail. 
In contrast, our model could actively use prior knowledge from past states, which ensures a high probability of recapturing target when it reappears. In the dense scene (on the top right), M$^2$-Track capture a specific target with a relatively tight fit. However, benefiting from utilizing sequence data, our model can produce a trajectory that aligns more consistently with the ground truth. 
Different from the vehicles that typically maintain a certain distance from each other, the pedestrians are often close together and easily mislead tracker to the incorrect object (on the bottom right). Nevertheless, due to considering the historical states, our model could suppress sudden wrong shifts that do not conform to motion patterns and prevent the box from being erroneously assigned to another object.

\begin{table}[t]
\renewcommand{\arraystretch}{1.0}
\renewcommand\tabcolsep{8.0pt}
\centering
\caption{\textbf{Comparison of inference speed.}}
\label{tab:speed} 
\footnotesize
\begin{tabular}{c|cccc}
\toprule[.05cm]
Method & P2B       & BAT          & CXTrack       & PTTR          \\ \hline
\rule{0pt}{8pt}
FPS    & 45.5     & \textbf{57.0}         & 34.0          & \underline{51.0}            \\ \hline \hline \rule{0pt}{8pt}
Method & GLT-T    & M$^2$-Track      & STTracker       & \textbf{SeqTrack3D}       \\ \hline
\rule{0pt}{8pt}
FPS    &   30.0   & \textbf{57.0}    & 23.6            &   38.0       \\ \toprule[.05cm]
\end{tabular}
\vspace{-0.7cm}
\end{table}

\subsection{Ablation Study.}
In this section, to thoroughly explore the effect of different components and settings in the proposed SeqTrack3D, we perform a series of ablation studies on Car and Pedestrian categories of NuScenes, which are the predominant classes. Specifically, these experiments mainly focus on three aspects: the contributions of individual model components, the impact of varying time window length, and the influence of explicitly constraining predicted historical BBoxes during training. For evaluation purposes, we train on 1/5 of the training set and assess the complete validation set.

 \noindent\textbf{Model Components.} Our SeqTrack3D mainly consists of an encoder and decoder. In order to investigate the interplay between them and comprehend their importance in overall performance, we conduct experiments focusing on each model component. In Table.~\ref{tab:ablation}, when using optimal time window length $N=4$, we observe that omitting the local and global features will significantly damage model performance. 
Then, a moderate progress is noticed when only global encoder is employed to capture inter-frame dynamics. It is also worth noting that solely using local encoder would be better than global one. We speculate that the intra-frame geometry cue is important for accurate bounding box prediction. 
Finally, combining global and local information can achieve optimal result, which proves that decoupling local and global feature extraction allows the tracker to learn both geometric and dynamic aspects of target in point sequence.

\begin{table}[t]
\renewcommand{\arraystretch}{1.0}
\renewcommand\tabcolsep{8.0pt}
  \begin{center}
  \footnotesize
      \caption{\textbf{Ablation study of components in SeqTrack3D.}
       ``C" is Coarse Box Prediction. ``L" and ``G" refer to Local and Global Encoder. ``D" means Decoder.}~\label{tab:ablation}
       \vspace{-0.1cm}
    \begin{tabular}{c|cc|cc}
    \toprule[.05cm]
    \multirow{2}{*}{Component} & \multicolumn{2}{c|}{Car} & \multicolumn{2}{c}{Pedestrian} \\
     & Success & Precision & Success & Precision \\
    \hline \hline \rule{0pt}{8pt}
    C         & 51.42   & 59.66      & 30.89      & 57.65       \\
    C + G + D & 52.17   & 62.60      & 31.83      & 59.06       \\
    C + L + D & 54.30   & 64.57      & 32.05      & 59.22       \\
    C + L + G + D & \textbf{56.73} & \textbf{66.01} & \textbf{32.80} & \textbf{61.53} \\
    \toprule[.05cm]
    \end{tabular}
    \end{center}
    \vspace{-0.4cm}
\end{table}%

\begin{table}[t]
\renewcommand{\arraystretch}{1.0}
\renewcommand\tabcolsep{8.0pt}
  \centering
  \footnotesize
  \caption{\textbf{Ablation study on the time window length.}}
  \vspace{-0.1cm}
  \begin{tabular}{c|cc|cc}
  \toprule[.05cm]
  \multirow{2}{*}{Time Length $N$} & \multicolumn{2}{c|}{Car} & \multicolumn{2}{c}{Pedestrian} \\
  & Success & Precision & Success & Precision \\
  \hline \hline \rule{0pt}{8pt}
  1 + 1        & 51.60 & 61.30 & 31.94 & 60.57 \\
  1 + 3 (Ours)       & \textbf{56.73} & \textbf{66.01} & 32.80 & 61.53 \\
  1 + 5        & 54.57 & 62.98 & 32.48 & \textbf{62.72} \\
  1 + 7        & 51.15 & 60.07 & \textbf{33.44} & 62.70 \\
  \toprule[.05cm]
  \end{tabular}
  \label{tab:framenumber}
  \vspace{-0.4cm}
\end{table}
\noindent\textbf{Time Window Length.} The time window length controls how far back in time our model draws knowledge from. It's intuitive to consider that earlier historical frames contain older states of a target, but too old frames might not always contribute positively to the prediction in the current frame. Additionally, longer time windows during training can make it challenging to simulate conditions in the testing phase. Based on these considerations, we conduct experiments on the time window length to determine the optimal setting for SeqTrack3D. As illustrated in Table.~\ref{tab:framenumber}, the Car improves \textit{Success} by 5.13\% and \textit{Precision} by 4.71\% when time length increases from 1+1 to 1+3, indicating the positive impact of incorporating historical data.
However, when the number of frames increases to 1+7, the tracking performance almost degrades to be equivalent to 1+1 frames. 
We believe that the reasons are two-fold: First, the random offsets for longer historical BBox sequences during training might make it difficult to simulate test conditions; Second, more cumulative errors contained in the historical boxes will also contribute to performance drop.

\begin{table}[t]
\renewcommand{\arraystretch}{1.0}
\renewcommand\tabcolsep{7.0pt}
  \begin{center}
  \footnotesize
  \caption{\textbf{Ablation study on the object-wise constraints for box sequence.} X+Y indicates constraints applied to X current and Y previous boxes.}
  \label{tab:constriant}
  \vspace{-0.1cm}
  \begin{tabular}{>{\centering\arraybackslash}p{2cm}|cc|cc}  
  \toprule[.05cm]
  \multirow{2}{*}{Constraints Length} & \multicolumn{2}{c|}{Car} & \multicolumn{2}{c}{Pedestrian} \\
  & Success & Precision & Success & Precision \\
  \hline \hline \rule{0pt}{8pt}
  1 + 0       & 53.09 & 63.03 & 30.17 & 58.62 \\
  1 + 1       & 55.22 & 64.58 & 30.52 & 59.21 \\
  1 + 2       & 55.83 & 65.40 & 31.12 & 59.49 \\
  1 + 3 (Ours)       & \textbf{56.73} & \textbf{66.01} & \textbf{32.80} & \textbf{61.53} \\
  \toprule[.05cm]
  \end{tabular}
  \end{center}
  \vspace{-0.6cm}
\end{table}

\noindent\textbf{Constraints Length.} Even though using BBox in the current frame as the final output, we still compute the loss of BBox in historical frame to impose constraints. This encourages the model to learn the motion patterns of target from previous priors. As demonstrated in Table.~\ref{tab:constriant}, we change the number of constraints on historical boxes while fixing time window length at \( N=4 \). The results indicate when the number of constraints equals to the sequence length, our method reaches optimal performance.

\section{CONCLUSIONS}

In this paper, we propose a Seq2Seq tracking framework to address 3D SOT task. By leveraging Transformer to process sequence data, our SeqTrack3D could capture dynamic traits of target from point cloud and BBox sequences and result in a remarkable performance improvement. 
However, we notice a degradation as the length of point sequence increases.
In future work, we will further explore methods to effectively capture spatial and temporal information in long sequences.

\bibliographystyle{ieeetr} 
\bibliography{ref}

\end{document}